\documentclass[sigconf,natbib=true]{acmart}

\AtBeginDocument{%
  \providecommand\BibTeX{{%
    \normalfont B\kern-0.5em{\scshape i\kern-0.25em b}\kern-0.8em\TeX}}}

\setcopyright{acmcopyright}
\copyrightyear{2023}
\acmYear{2023}
\acmDOI{XXXXXXX.XXXXXXX}

\acmConference[CIKM'23]{Make sure to enter the correct
  conference title from your rights confirmation email}{Oct 21--25,
  2023}{Birmingham, UK}
\acmPrice{15.00}
\acmISBN{978-1-4503-XXXX-X/18/06}

\acmSubmissionID{2469}

\usepackage{algorithmic}
\usepackage{graphicx, multirow}
\usepackage{textcomp}
\usepackage{xcolor}
\usepackage{comment}

\begin{document}

\title{SmartFlow: Robotic Process Automation using LLMs}


\author{Arushi Jain, Shubham Paliwal, Monika Sharma, Lovekesh Vig, Gautam Shroff}
\affiliation{%
  \institution{TCS Research}
  \city{New Delhi}
  \country{India}}
\email{(j.arushi, shubham.p3, monika.sharma1, lovekesh.vig, gautam.shroff)@tcs.com}

\renewcommand{\shortauthors}{Arushi et al.}

\begin{abstract}
Robotic Process Automation (RPA) systems face challenges in handling complex processes and diverse screen layouts that require advanced human-like decision-making capabilities. These systems typically rely on pixel-level encoding through drag-and-drop or automation frameworks such as Selenium to create navigation workflows, rather than visual understanding of screen elements. In this context, we present SmartFlow, an AI-based RPA system that uses pre-trained large language models (LLMs) coupled with deep-learning based image understanding. Our system can adapt to new scenarios, including changes in the user interface and variations in input data, without the need for human intervention. SmartFlow uses computer vision and natural language processing to perceive visible elements on the graphical user interface (GUI) and convert them into a textual representation. This information is then utilized by LLMs to generate a sequence of actions that are executed by a scripting engine to complete an assigned task. To assess the effectiveness of SmartFlow, we have developed a dataset that includes a set of generic enterprise applications with diverse layouts, which we are releasing for research use. Our evaluations on this dataset demonstrate that SmartFlow exhibits robustness across different layouts and applications. SmartFlow can automate a wide range of business processes such as form filling, customer service, invoice processing, and back-office operations. SmartFlow can thus assist organizations in enhancing productivity by automating an even larger fraction of screen-based workflows. The demo-video and dataset are available at \url{https://smartflow-4c5a0a.webflow.io/}.

\end{abstract}



\maketitle

\section{Introduction}
\label{sec:intro}
Robotic Process Automation (RPA)~\cite{rpa1} has garnered substantial interest as a means of automating repetitive and labor-intensive business processes through software bots. Its adoption spans various industries, including customer service, finance, human resources, supply chain management, and healthcare with the aim of enhancing operational efficiency, minimizing costs and errors, and improving overall customer experience~\cite{rpa-service1, rpa-service2, rpa-banking1, rpa-banking2, rpa-hr1, rpa-hr2, rpa-supply, rpa-health1}.
Despite its popularity, scientific literature on RPA is limited, with existing sources mainly focusing on its features and benefits~\cite{rpa1, rpa2, rpa-review, rpa3, rpa4, rpa5}. Current RPA systems have inherent limitations concerning decision-making, language comprehension, and visual capabilities, as they are designed to adhere to pre-defined rules and workflows using pixel-level encoding of the graphical user interface (GUI)\cite{uipath, automation-anywhere, blueprism}. These functionalities are typically implemented through drag-and-drop interfaces, screenplay recording, or automation frameworks such as Selenium\cite{selenium}. Consequently, these systems lack flexibility in adapting to changes in the UI and struggle to handle tasks that require intricate visual analysis and natural language understanding.

Recent years have witnessed remarkable progress in deep learning and computer vision, leading to advancements in object recognition, image segmentation, and video analysis~\cite{detr, segment-anything, video-analysis}. 
Additionally, the introduction of pre-trained large language models such as GPT-3~\cite{gpt3}, ChatGPT~\cite{chatgpt}, Llama~\cite{llama}, and PaLM~\cite{palm} has revolutionized natural language processing, enabling advanced language understanding and generation capabilities. For instance, AI agents such as AgentGPT~\footnote{AgentGPT: https://github.com/reworkd/AgentGPT} and AutoGPT~\footnote{AutoGPT: https://github.com/Significant-Gravitas/Auto-GPT} can automate a wide range of tasks, including writing, translation, and content generation. Moreover, the advent of Visual Language Models (VLMs) such as Control-Net~\cite{control-net} and Visual-ChatGPT~\cite{visual-chatgpt}, combining text-based LLMs with visual understanding, has opened new avenues for image analysis and processing. While VLMs such as Visual-ChatGPT~\cite{visual-chatgpt} and Google's Bard~\footnote{Google's Bard: https://bard.google.com/} can perform tasks such as generating images from textual input, providing image descriptions, and answering questions about images, they require fine-tuning on Web GUIs datasets to identify and localize screen elements in application GUIs. Further, the recently announced GPT-4~\cite{gpt4} by OpenAI has received significant attention due to its promising capabilities in handling multimodal data. However, as of now, GPT-4 has not been available to everyone publicly, and its utility and limitations in handling visual data are yet to be evaluated.

\begin{figure*}[]
  \centering
  \includegraphics[width=0.8\textwidth]{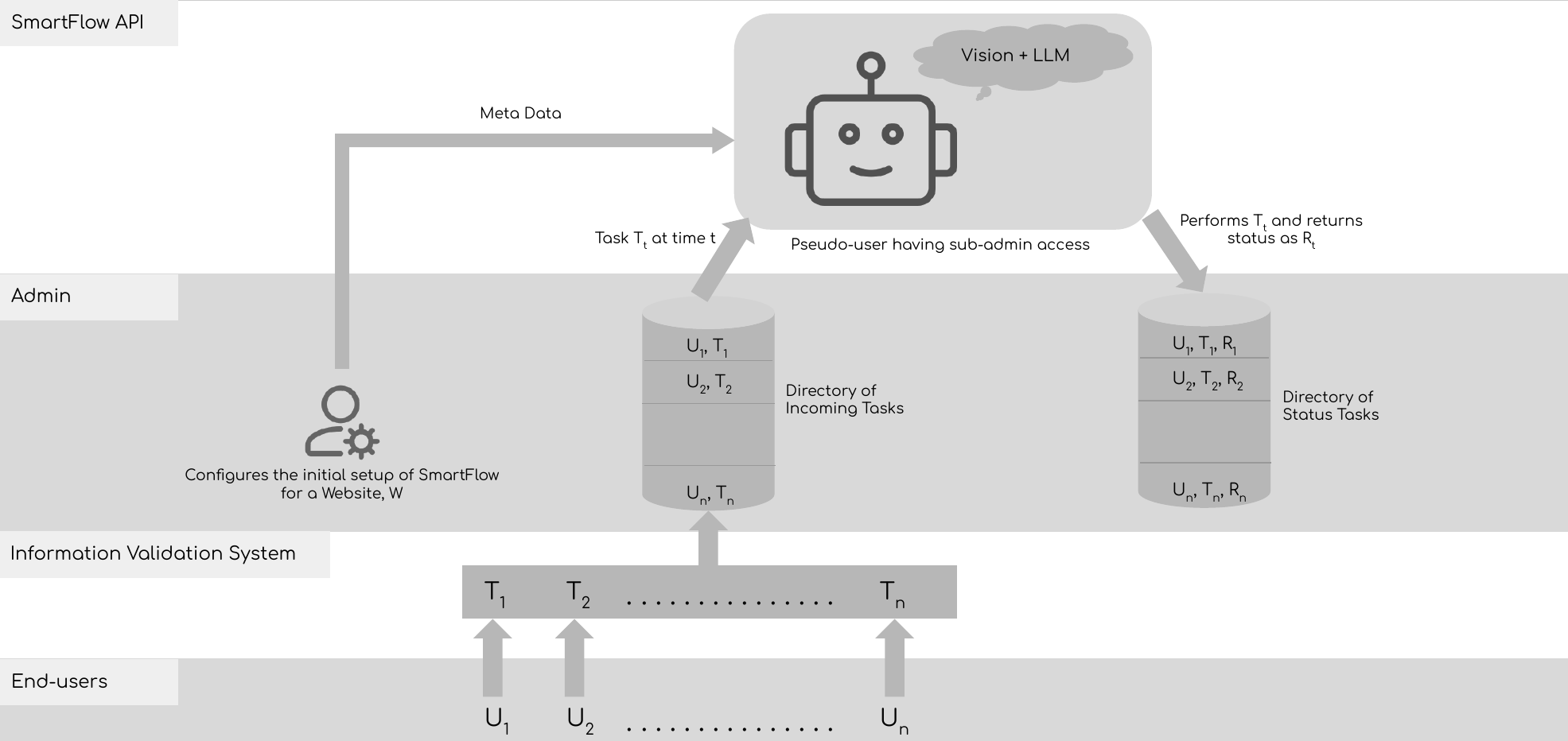}
  \caption{\small{The proposed system design of SmartFlow RPA: The process begins with end users ($U_1$, $U_2$,...,$U_n$) submitting corresponding task-requests ($T_1$, $T_2$,...,$T_n$) to fill out a form in a web application, which must include all mandatory details. These requests are then validated by an Information Validation System (IVS) and are added to a directory of incoming tasks for processing by the Smartflow system. As a one-time setup, the administrator specifies the URL of the website where end user requests are executed, along with the required HTML source code of the application pages and corresponding layout mapping of the input fields. The SmartFlow executes each task-request using machine vision and natural language understanding techniques, returning the status of the completed tasks to the directory of task status. }}
  \label{fig:smartflow-overview}
\end{figure*}

These breakthroughs have opened up new possibilities for integrating LLMs with RPA systems towards enabling them to perceive and autonomously interact with complex web applications. For example, Wang et al.\cite{mobileui-llm} conducted a study exploring the use of pre-trained language models (LLMs) to enable conversational interaction on mobile user interfaces (UIs). Their research involved providing GUIs to LLMs that were pre-trained for natural language understanding, along with employing various techniques to prompt the LLMs to perform conversational tasks. In another study\cite{rpa-ml}, Pedro et al. utilized the Yolo object detector~\cite{yolo} to identify screen elements such as menus and buttons. However, the study did not propose a method for determining the necessary actions to complete a specific task based on the identified screen elements. Additionally, the training of the object detector was limited to detecting Eclipse IDE screen elements only, requiring the development of a new detector in case of changes in the application type.

To address the limitations of current RPA systems, we propose a novel AI-based RPA system called SmartFlow that uses LLMs coupled with deep-learning based image understanding. It integrates vision capabilities with natural language processing techniques to adapt to changes in the graphical user interface (GUI) and automatically generate navigation workflows. By utilizing vision techniques, SmartFlow identifies and locates screen elements, while the HTML source code provides information about the type of these elements. A pre-trained large language model such as GPT-3 is then employed to generate navigation workflows based on this information. This navigation workflow is then executed using a scripting language to complete the assigned task. One notable benefit of SmartFlow is its ability to handle diverse application layouts and screen resolutions efficiently. In summary, our paper presents the following contributions:
\vspace{-2mm}
\begin{itemize}
    \item We propose an AI-based RPA system called SmartFlow which utilizes pre-trained LLMs in tandem with deep-vision and is capable of autonomously executing user-assigned tasks.
    \item SmartFlow leverages HTML code, visual and natural language understanding to interpret the layout mapping. This includes associating field names, their types, and corresponding placeholders/edit fields.
    \item SmartFlow is designed to be adaptable to GUI changes and handle complex tasks effectively. It achieves this by generating navigation workflows using vision and large language models (LLMs), without relying on predefined pixel-encoded rule-based workflows
    \item We demonstrate SmartFlow's proficiency in handling multi-page form submission applications with diverse field types, such as date pickers, dropdown menus, etc. through the use of vision-based algorithms. 
    \item To demonstrate the effectiveness of SmartFlow, we have curated a dataset called RPA-Dataset, containing generic web applications with various layouts. We have released this dataset publicly to foster research in this field.
\end{itemize}

\section{Overview}
\label{sec:prob-statement}
\begin{figure}[!htbp]
  \centering
  \includegraphics[width=0.45\textwidth]{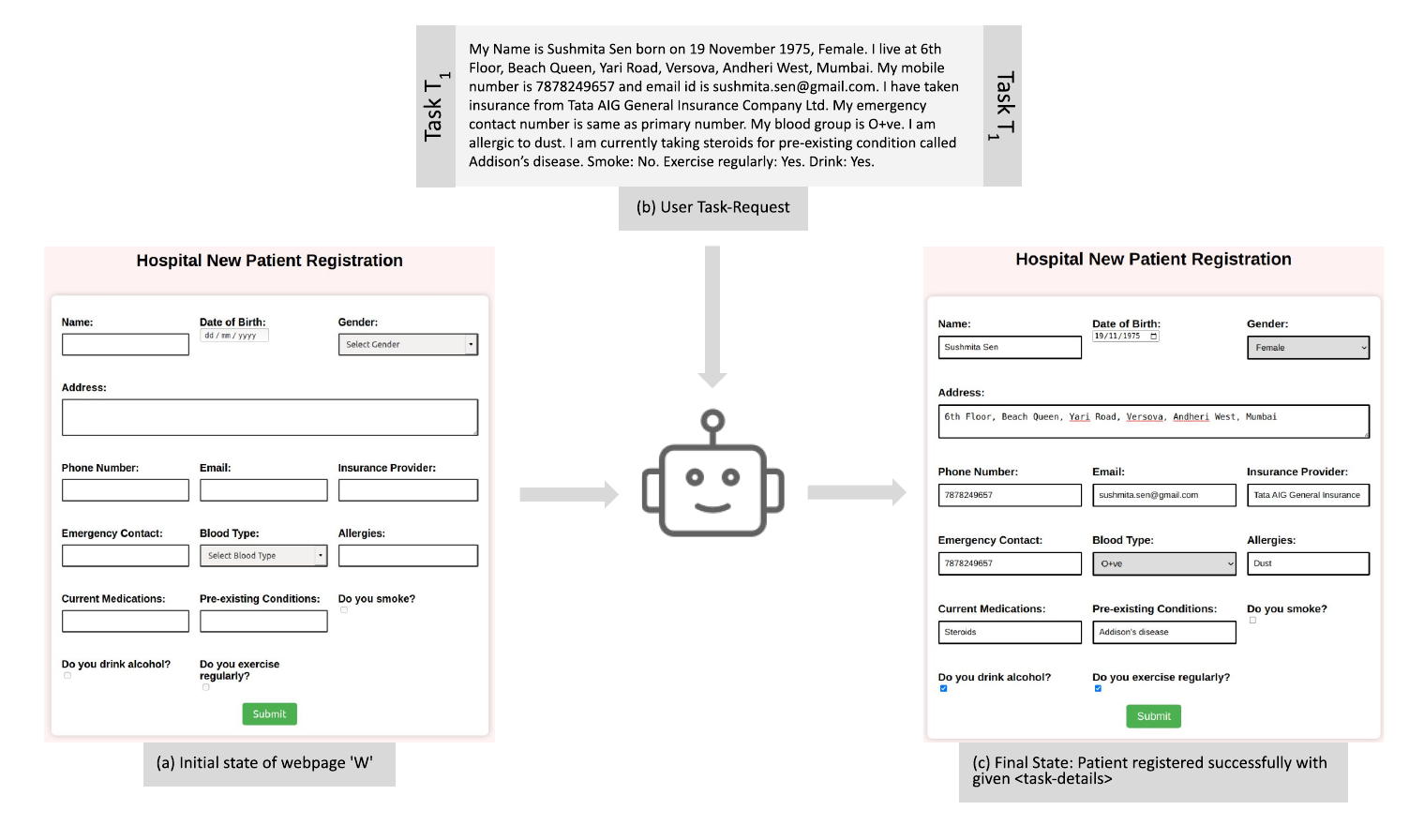}
  \caption{Given a user task-request to register a new patient on the hospital website $W$ with given details as $T_1$, SmartFlow performs the task automatically and returns the status as "Patient registered successfully".}
  \label{fig:prob-statement}
\end{figure}

Our objective is to automate the generation of navigation workflows for specific tasks within a graphical user interface (GUI) application. Using deep-vision and natural language understanding, we identify screen elements such as field names, placeholders/edit fields, and hints. Subsequently, LLM is used to determine the necessary actions to fill in the required information which are then executed using a scripting engine. Finally, SmartFlow provides updates on the status of the executed task. For example, let's consider the task of registering a new patient in a Hospital system, as shown in Figure~\ref{fig:prob-statement}. Traditionally, this process involved manual data entry from a handwritten document. However, we can automate it by digitizing the document using information extraction techniques~\cite{attend, deepreader, pick}. SmartFlow then automatically fills in the patient's details in the registration system, eliminating the need for manual data entry.

\section{System Design: SmartFlow}
\label{sec:proposed-system}
In Figure~\ref{fig:smartflow-overview}, we present an overview of our proposed RPA system. We envisage the following four user-classes for SmartFlow:
\vspace{-2mm}
\begin{itemize}
 \item \textit{End-user}: provides all the necessary information, such as task request data, via email or chat-bot to the application, with the objective of having the task executed automatically. 
 \item \textit{Information Validation System (IVS)}: ensures that the task-request received from the end-user is complete and includes all the required information for filling in the data fields necessary to complete the task and adds the task-request to the incoming task directory.
 \item \textit{Admin}: is responsible for setting up and configuring SmartFlow initially for an application. This involves providing meta-data such as the website URL and HTML source code for all its pages. The Admin also performs layout mapping, which associates visible field names on the application screen with their respective edit-fields and data-hints. We propose two vision-based methods for automatic layout mapping, which are validated by the Admin. In case of any errors, a demonstration approach is used for accurate layout mapping.
 \item \textit{SmartFlow API}: is responsible for sequentially handling task requests from the incoming requests directory. Upon completion, the API returns the task's output status (e.g., success, failure, or errors) to the task-status directory. The Admin is then responsible for communicating the task statuses to end-users through their preferred communication channel.
\end{itemize}

\begin{figure*}[t]
  \centering
  \includegraphics[width=1\textwidth]{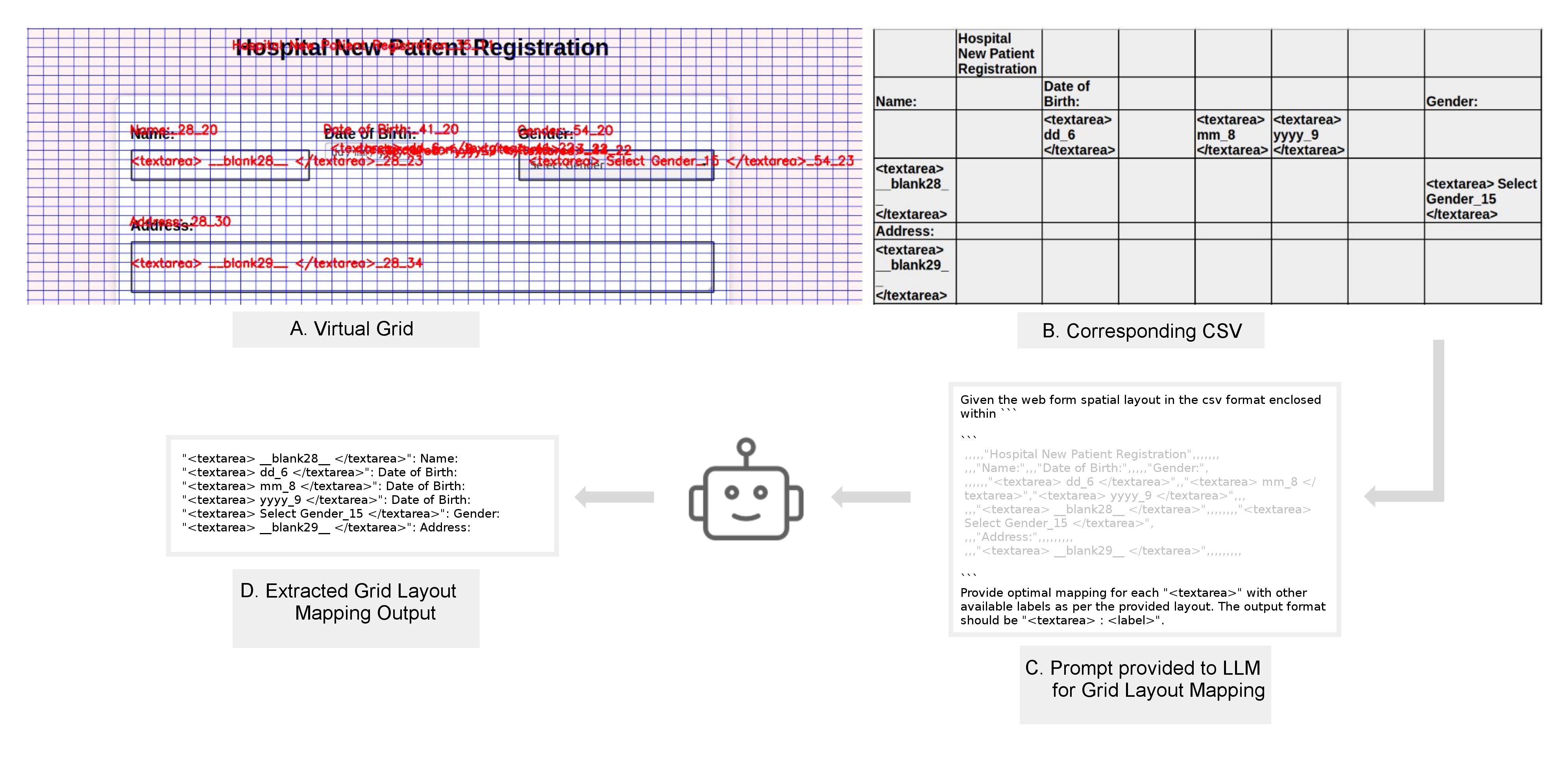}
  \vspace{-7mm}
  \caption{The Layout Mapping using Virtual Grid approach condenses the original layout by converting pixel coordinates into a virtual grid space (A) which is represented in a corresponding CSV format (B). The red colored text in (A) represents the OCR output along with corresponding grid coordinates. Next, the CSV file is fed into LLM as input with a prompt to give the mapping of field names with the corresponding edit-fields and data-hints (C). The output is the grid layout mapping (D).}
  \label{fig:layout-grid}
\end{figure*}

Next, we will present three different methods for \textbf{Layout Mapping}, each offering unique advantages.
\begin{enumerate}
\item \textbf{Rule-Based Approach:} After analyzing multiple web application forms, we observed a consistent pattern where field names are usually aligned to the left or top of the edit field, while data-hints are commonly positioned at the bottom or right side. Leveraging these observations, we have devised an automated layout mapping technique that combines vision-based methods with predefined rules and heuristics.

\item \textbf{Virtual Grid Approach:} Typically, Layout Mapping Models face challenges in interpreting pixel coordinates to comprehend spatial layouts accurately. Hence, we propose condensing the original layout by converting pixel coordinates into a virtual grid space, as shown in Figure~\ref{fig:layout-grid}. Each unit in the virtual grid covers multiple pixel blocks, simplifying misalignment checks to eight neighboring cells and reducing the spatial complexity. We represent the spatial layout using .CSV format in virtual grids, which are fed as input to LLM along with a text prompt to generate the layout mapping.

\item \textbf{Demonstration by Admin:} If a rule-based or virtual grid-based approach does not yield accurate layout mapping, a demonstration-based approach can be employed. This method involves the administrator providing a demonstration by entering dummy data into the web application form and submitting a JSON file with relevant field information. SmartFlow uses visual analysis of the filled and unfilled form images, along with the JSON data, to establish connections between field names, placeholders, and values, ensuring $100\%$ accurate layout mapping.
\end{enumerate}

\begin{figure*}[t]
  \centering
  \includegraphics[width=0.95\textwidth]{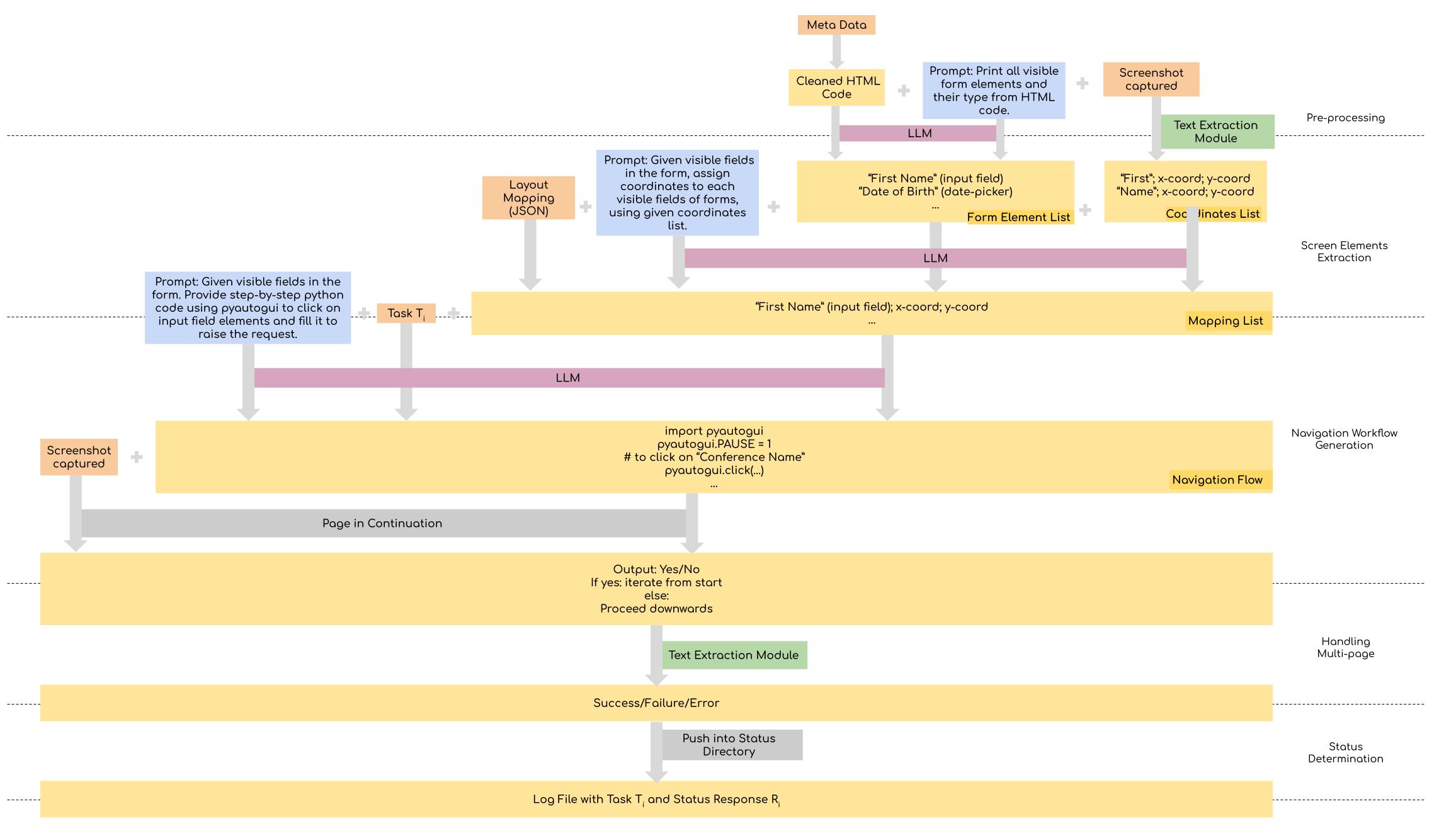}
  \vspace{-2mm}
  \caption{\small{Pipeline of Smartflow}}
  \label{fig:pipeline-smartflow}
\end{figure*} 

\noindent \textbf{SmartFlow Algorithm}: The Smartflow, as shown in Figure~\ref{fig:pipeline-smartflow}, begins processing by sequentially extracting and handling requests from the incoming request queue. 

\begin{itemize}
    \item \textit{Pre-processing}: It cleans the HTML source code provided in the input metadata (application URL, HTML source code and layout mapping), ensuring it meets the size limit of Large Language Models. This involves removing unnecessary attributes and classes from the HTML source code that could hinder LLM processing.
    
    \item \textit{Form Elements Extraction}: The cleaned HTML source code is input into pre-trained Large Language Models (LLMs) such as GPT-3~\cite{gpt3} and ChatGPT~\cite{chatgpt} to extract field names and types using the prompt as shown in Figure~\ref{fig:pipeline-smartflow}. Meanwhile, a screenshot image of the application is captured, and text-regions are extracted using EasyOCR\footnote{EasyOCR: https://github.com/JaidedAI/EasyOCR}. The extracted information is merged with the layout mapping by giving a text-prompt to LLM to create a Mapping List, which includes field names, types, and coordinates. This Mapping List serves as the textual representation of the visual screen for generating the navigation workflow.
    

    \item \textit{Navigation Workflow Generation}: In this step, the Mapping List and task-request are given as input to the LLM with a prompt to generate PyAutoGUI~\footnote{PyAutoGUI Documentation: https://pyautogui.readthedocs.io/en/latest/} code. This scripting code determines the sequence of actions, including clicking on the correct form-field, to complete the task-request accurately. The precision is crucial to avoid incorrect form submission. The algorithm executes micro-level steps with high accuracy and handles different field types such as date pickers, dropdown menus, radio buttons, and checkboxes using vision-based algorithms. The details of these algorithms are provided as follows:

    \begin{itemize}
    \item \textit{Datepickers}: They are characterized by their diverse and captivating designs, and pose a unique challenge in the realm of user interaction. In the majority of the layouts encountered, datepickers can be effortlessly populated by simply typing the relevant date. However, in scenarios where manual input is restricted, SmartFlow employs an astute strategy to ensure optimal performance. It intelligently scrolls through the corresponding year section by using PyAutoGUI and visually analyzing the screenshot of the calendar and hence, diligently seeks to match the desired year, adapting its direction based on whether the year lies in the past or future. It further refines the selection process by meticulously clicking through each subsequent month until the desired month aligns harmoniously. With the calendar now seamlessly set, Smartflow harnesses the power of vision and a simple technique of the calendar calculation to extrapolate the coordinate for each date, thus enabling flawless and accurate selection.
    
    \item \textit{Dropdown}: SmartFlow, our proposed RPA system, effectively handles complex dropdown menus that have initially hidden options by leveraging visual processing techniques. When it encounters a <select> field, it clicks on the designated area of a dropdown menu, captures a screenshot to eliminate visual discrepancies, and extracts the dropdown options as digital representations using a text extraction module. In cases where the desired selection is not immediately visible, SmartFlow intelligently scrolls the dropdown panel until it locates the target selection using visual analysis of captured screen, ensuring a smooth and seamless user experience.

    \item \textit{Radio-buttons / Checkboxes}: To handle radio-buttons / checkboxes, Smartflow recognizes that it doesn't require clicking directly on the checkbox or circle. It utilizes the text associated with the option for selection. Smartflow interacts with ChatGPT to gather the available options and coordinates, and a dedicated module performs the action of clicking on the desired option from the list.
    \end{itemize} 
    
    These algorithms for handling complex fields are invoked by the LLM during the generation of the navigation workflow. They are premised on a comprehensive understanding of the visual aspects of the form and integrate nuanced insights into the behavioral patterns of each field type.

    \item \textit{Handling Multi-page Form Submission}: After executing the navigation workflow using the scripting code, SmartFlow captures another screenshot to handle multi-page forms effectively. By leveraging visual cues from the website's layout, it recognizes the continuation of the form and sequentially processes the user's requests to fill in any remaining fields.

    \item \textit{Determining the status of executed task-requests}: SmartFlow employs a frame difference technique to extract feedback messages related to the success, failure, or errors encountered during form submission. These messages, obtained using a text-extractor, can address network connectivity, missing fields, or successful submissions. By logging these messages into a status queue, SmartFlow facilitates analysis and improves the user experience.
\end{itemize}

\vspace{-2mm}
\section{Dataset Details}
\label{sec:dataset}
To evaluate the effectiveness of integrating vision and large language models (LLMs) in RPA systems, we created the RPA-Dataset and released it for public use\footnote{RPA-Dataset: \url{https://drive.google.com/file/d/1ddTfEFLUeIIzekLANr_4yMNWzdYs-3Ls/view}}. For this dataset, we identified five distinct applications that exhibit a substantial demand for RPA integration and developed different HTML websites. These applications span ubiquitous enterprise domains, including but not limited to a Conference Attendance System, New Patient Registration, Sales Lead Generation, Customer Complaint Handling, and Passport Registration. Each website has a maximum of five variations in layouts which showcase variations in design and development approaches. Each website layout has five user-task requests. The RPA-Dataset includes the source HTML codes of the applications, along with ground-truth annotations for tasks such as OCR (Optical Character Recognition), Layout Mapping, filling data fields, and handling complex fields such as dropdowns, datepickers and radio-buttons/checkboxes. The preparatory task files for input, along with the corresponding filled ground truth data, have been manually generated and organized in CSV format.

\begin{table*}[]
\centering
\caption{\small{Table showing the performance of SmartFlow on CAS application.}}
\label{tab:results-cas}
\vspace{-2mm}
\resizebox{0.85\textwidth}{!}{%
\begin{tabular}{|c|c|cccccc|c|ccc|}
\hline
\multirow{3}{*}{\textbf{Layout No.}} & \multirow{3}{*}{\textbf{Page No.}} & \multicolumn{6}{c|}{\textbf{Accuracy}}                                                                                                                                                                                                                                        & \multirow{3}{*}{\textbf{\begin{tabular}[c]{@{}c@{}}Task completion average time\\ (in mins)\end{tabular}}} & \multicolumn{3}{c|}{\textbf{Complex Component Accuracy}}                                       \\ \cline{3-8} \cline{10-12} 
                                     &                                    & \multicolumn{2}{c|}{\textbf{OCR}}                                         & \multicolumn{2}{c|}{\textbf{Layout Mapping}}                                          & \multicolumn{1}{c|}{\multirow{2}{*}{\textbf{Filled Data}}} & \multirow{2}{*}{\textbf{Request Submission}} &                                                                                                            & \multicolumn{1}{c|}{\multirow{2}{*}{\textbf{Datepicker}}} & \multicolumn{1}{c|}{\multirow{2}{*}{\textbf{Dropdown}}} &
                                     \multirow{2}{*}{\textbf{Radio/Checkbox}}\\ \cline{3-6}
                                     &                                    & \multicolumn{1}{c|}{\textbf{CER}}   & \multicolumn{1}{c|}{\textbf{WER}}   & \multicolumn{1}{c|}{\textbf{Rule-based}} & \multicolumn{1}{c|}{\textbf{Virtual-Grid}} & \multicolumn{1}{c|}{}                                      &                                              &                                                                                                            & \multicolumn{1}{c|}{}                                     & \multicolumn{1}{c|}{}
                                     & \\ \hline
\multirow{2}{*}{1}                   & 1                                  & \multicolumn{1}{c|}{0.005}          & \multicolumn{1}{c|}{0.087}          & \multicolumn{1}{c|}{1.0}                 & \multicolumn{1}{c|}{0.91}                      & \multicolumn{1}{c|}{0.91}                                  & \multirow{2}{*}{1.0}                         & \multirow{2}{*}{4.27}                                                                                      & \multicolumn{1}{c|}{1.0}                                  & \multicolumn{1}{c|}{1.0} & 0.8                                \\ \cline{2-7} \cline{10-12} 
                                     & 2                                  & \multicolumn{1}{c|}{0.004}          & \multicolumn{1}{c|}{0.050}          & \multicolumn{1}{c|}{1.0}                 & \multicolumn{1}{c|}{1.0}                      & \multicolumn{1}{c|}{1.0}                                   &                                              &                                                                                                            & \multicolumn{1}{c|}{}                                     & 
                                     \multicolumn{1}{c|}{}                                      &\\ \hline
\multirow{2}{*}{2}                   & 1                                  & \multicolumn{1}{c|}{0.045}          & \multicolumn{1}{c|}{0.176}          & \multicolumn{1}{c|}{0.8}                 & \multicolumn{1}{c|}{0.91}                      & \multicolumn{1}{c|}{0.91}                                  & \multirow{2}{*}{1.0}                         & \multirow{2}{*}{6.4}                                                                                       & \multicolumn{1}{c|}{1.0}                                  & \multicolumn{1}{c|}{1.0} & 0.8                                \\ \cline{2-7} \cline{10-12} 
                                     & 2                                  & \multicolumn{1}{c|}{0.004}          & \multicolumn{1}{c|}{0.025}          & \multicolumn{1}{c|}{1.0}                 & \multicolumn{1}{c|}{1.0}                      & \multicolumn{1}{c|}{1.0}                                   &                                              &                                                                                                            & \multicolumn{1}{c|}{}                                     &  
                                     \multicolumn{1}{c|}{}                                  &\\ \hline
\multirow{2}{*}{3}                   & 1                                  & \multicolumn{1}{c|}{0.037}          & \multicolumn{1}{c|}{0.154}          & \multicolumn{1}{c|}{0.9}                 & \multicolumn{1}{c|}{0.91}                      & \multicolumn{1}{c|}{0.91}                                  & \multirow{2}{*}{1.0}                         & \multirow{2}{*}{6.7}                                                                                       & \multicolumn{1}{c|}{0.933}                                & \multicolumn{1}{c|}{1.0} & 0.8                                \\ \cline{2-7} \cline{10-12} 
                                     & 2                                  & \multicolumn{1}{c|}{0.008}          & \multicolumn{1}{c|}{0.075}          & \multicolumn{1}{c|}{1.0}                 & \multicolumn{1}{c|}{1.0}                      & \multicolumn{1}{c|}{1.0}                                   &                                              &                                                                                                            & \multicolumn{1}{c|}{}                                     &
                                     \multicolumn{1}{c|}{}                                  &\\ \hline
\multirow{2}{*}{4}                   & 1                                  & \multicolumn{1}{c|}{0.016}          & \multicolumn{1}{c|}{0.091}          & \multicolumn{1}{c|}{1.0}                 & \multicolumn{1}{c|}{0.91}                      & \multicolumn{1}{c|}{0.91}                                  & \multirow{2}{*}{1.0}                         & \multirow{2}{*}{6.8}                                                                                       & \multicolumn{1}{c|}{1.0}                                  & \multicolumn{1}{c|}{1.0} & 0.8                                \\ \cline{2-7} \cline{10-12} 
                                     & 2                                  & \multicolumn{1}{c|}{0.005}          & \multicolumn{1}{c|}{0.054}          & \multicolumn{1}{c|}{1.0}                 & \multicolumn{1}{c|}{1.0}                      & \multicolumn{1}{c|}{1.0}                                   &                                              &                                                                                                            & \multicolumn{1}{c|}{}                                     &
                                     \multicolumn{1}{c|}{}                                  &\\ \hline
\multirow{2}{*}{5}                   & 1                                  & \multicolumn{1}{c|}{0.032}          & \multicolumn{1}{c|}{0.107}          & \multicolumn{1}{c|}{1.0}                 & \multicolumn{1}{c|}{0.91}                      & \multicolumn{1}{c|}{0.98}                                  & \multirow{2}{*}{1.0}                         & \multirow{2}{*}{4.8}                                                                                       & \multicolumn{1}{c|}{1.0}                                  & \multicolumn{1}{c|}{1.0} & 0.0                                \\ \cline{2-7} \cline{10-12} 
                                     & 2                                  & \multicolumn{1}{c|}{0.005}          & \multicolumn{1}{c|}{0.027}          & \multicolumn{1}{c|}{1.0}                 & \multicolumn{1}{c|}{1.0}                      & \multicolumn{1}{c|}{1.0}                                   &                                              &                                                                                                            & \multicolumn{1}{c|}{}                                     &
                                     \multicolumn{1}{c|}{}                                  & \\ \hline
\textbf{Average}                     & \textbf{}                          & \multicolumn{1}{c|}{\textbf{0.015}} & \multicolumn{1}{c|}{\textbf{0.086}} & \multicolumn{1}{c|}{\textbf{0.97}}       & \multicolumn{1}{c|}{\textbf{0.955}}                      & \multicolumn{1}{c|}{\textbf{0.95}}                         & \textbf{1.0}                                 & \textbf{5.7}                                                                                               & \multicolumn{1}{c|}{\textbf{0.98}}                        & \multicolumn{1}{c|}{\textbf{1.0}} & \textbf{0.64}                       \\ \hline
\end{tabular}%
}
\end{table*}

\begin{table*}[]
\centering
\caption{\small{Table showing the performance of SmartFlow on Patient Registration application.}}
\label{tab:results-patient}
\resizebox{0.85\textwidth}{!}{%
\begin{tabular}{|c|c|cccccc|c|ccc|}
\hline
\multirow{3}{*}{\textbf{Layout No.}} &
  \multirow{3}{*}{\textbf{Page No.}} &
  \multicolumn{6}{c|}{\textbf{Accuracy}} &
  \multirow{3}{*}{\textbf{\begin{tabular}[c]{@{}c@{}}Task completion average time\\ (in mins)\end{tabular}}} &
  \multicolumn{3}{c|}{\textbf{Complex Component Accuracy}} \\ \cline{3-8} \cline{10-12} 
 &
   &
  \multicolumn{2}{c|}{\textbf{OCR}} &
  \multicolumn{2}{c|}{\textbf{Layout Mapping}} &
  \multicolumn{1}{c|}{\multirow{2}{*}{\textbf{Filled Data}}} &
  \multirow{2}{*}{\textbf{Request Submission}} &
   &
  \multicolumn{1}{c|}{\multirow{2}{*}{\textbf{Datepicker}}} &
  \multicolumn{1}{c|}{\multirow{2}{*}{\textbf{Dropdown}}} &
  \multirow{2}{*}{\textbf{Radio/Checkbox}} \\ \cline{3-6}
 &
   &
  \multicolumn{1}{c|}{\textbf{CER}} &
  \multicolumn{1}{c|}{\textbf{WER}} &
  \multicolumn{1}{c|}{\textbf{Rule-based}} &
  \multicolumn{1}{c|}{\textbf{Virtual-Grid}} &
  \multicolumn{1}{c|}{} &
   &
   &
  \multicolumn{1}{c|}{} &
  \multicolumn{1}{c|}{} &
   \\ \hline
1 &
  1 &
  \multicolumn{1}{c|}{0.0} &
  \multicolumn{1}{c|}{0.0} &
  \multicolumn{1}{c|}{0.8} &
  \multicolumn{1}{c|}{1.0} &
  \multicolumn{1}{c|}{0.96} &
  1.0 &
  1.99 &
  \multicolumn{1}{c|}{1.0} &
  \multicolumn{1}{c|}{1.0} &
  0.867 \\ \hline
2 &
  1 &
  \multicolumn{1}{c|}{0.0} &
  \multicolumn{1}{c|}{0.0} &
  \multicolumn{1}{c|}{1.0} &
  \multicolumn{1}{c|}{0.733} &
  \multicolumn{1}{c|}{0.91} &
  1.0 &
  1.59 &
  \multicolumn{1}{c|}{1.0} &
  \multicolumn{1}{c|}{1.0} &
  0.9 \\ \hline
3 &
  1 &
  \multicolumn{1}{c|}{0.0} &
  \multicolumn{1}{c|}{0.0} &
  \multicolumn{1}{c|}{0.8} &
  \multicolumn{1}{c|}{0.812} &
  \multicolumn{1}{c|}{0.96} &
  1.0 &
  2.29 &
  \multicolumn{1}{c|}{1.0} &
  \multicolumn{1}{c|}{1.0} &
  0.8 \\ \hline
4 &
  1 &
  \multicolumn{1}{c|}{0.0} &
  \multicolumn{1}{c|}{0.0} &
  \multicolumn{1}{c|}{1.0} &
  \multicolumn{1}{c|}{0.833} &
  \multicolumn{1}{c|}{0.96} &
  1.0 &
  2.16 &
  \multicolumn{1}{c|}{0.8} &
  \multicolumn{1}{c|}{1.0} &
  0.93 \\ \hline
5 &
  1 &
  \multicolumn{1}{c|}{0.0} &
  \multicolumn{1}{c|}{0.0} &
  \multicolumn{1}{c|}{1.0} &
  \multicolumn{1}{c|}{1.0} &
  \multicolumn{1}{c|}{0.97} &
  1.0 &
  1.73 &
  \multicolumn{1}{c|}{1.0} &
  \multicolumn{1}{c|}{1.0} &
  1.0 \\ \hline
\textbf{Average} &
  \textbf{} &
  \multicolumn{1}{c|}{\textbf{0.0}} &
  \multicolumn{1}{c|}{\textbf{0.0}} &
  \multicolumn{1}{c|}{\textbf{0.92}} &
  \multicolumn{1}{c|}{\textbf{0.876}} &
  \multicolumn{1}{c|}{\textbf{0.952}} &
  \textbf{1.0} &
  \textbf{1.952} &
  \multicolumn{1}{c|}{\textbf{0.96}} &
  \multicolumn{1}{c|}{\textbf{1.0}} &
  \textbf{0.88} \\ \hline
\end{tabular}%
}
\end{table*}

\begin{table*}[]
\centering
\caption{\small{Table showing the performance of SmartFlow across different applications with diverse layouts. We report average accuracy of all the layouts per application.}}
\label{tab:results-apps}
\vspace{-2mm}
\resizebox{0.85\textwidth}{!}{%
\begin{tabular}{|c|cccccc|c|ccc|}
\hline
\multirow{3}{*}{\textbf{Application}} & \multicolumn{6}{c|}{\textbf{Average Accuracy}}                                                                                                                                                                                                                                & \multirow{3}{*}{\textbf{\begin{tabular}[c]{@{}c@{}}Task completion average time\\ (in mins)\end{tabular}}} & \multicolumn{3}{c|}{\textbf{Complex Component Average Accuracy}}                               \\ \cline{2-7} \cline{9-11} 
                                      & \multicolumn{2}{c|}{\textbf{OCR}}                                         & \multicolumn{2}{c|}{\textbf{Layout Mapping}}                                          & \multicolumn{1}{c|}{\multirow{2}{*}{\textbf{Filled Data}}} & \multirow{2}{*}{\textbf{Request Submission}} &                                                                                                            & \multicolumn{1}{c|}{\multirow{2}{*}{\textbf{Datepicker}}} &
                                      \multicolumn{1}{c|}{\multirow{2}{*}{\textbf{Dropdown}}}
                                      & {\multirow{2}{*}{\textbf{Radio/Checkbox}}} \\ \cline{2-5}
                                      & \multicolumn{1}{c|}{\textbf{CER}}   & \multicolumn{1}{c|}{\textbf{WER}}   & \multicolumn{1}{c|}{\textbf{Rule-based}} & \multicolumn{1}{c|}{\textbf{Virtual-Grid}} & \multicolumn{1}{c|}{}                                      &                                              &                                                                                                            & \multicolumn{1}{c|}{}                                     & \multicolumn{1}{c|}{}                      &                                   \\ \hline
CAS                                   & \multicolumn{1}{c|}{0.015}          & \multicolumn{1}{c|}{0.086}          & \multicolumn{1}{c|}{0.97}                & \multicolumn{1}{c|}{0.955}                      & \multicolumn{1}{c|}{0.95}                                  & 1.0                                          & 5.7                                                                                                        & \multicolumn{1}{c|}{0.98}                                 & \multicolumn{1}{c|}{1.0} & 0.64                                \\ \hline
Patient Registration                  & \multicolumn{1}{c|}{0.0}          & \multicolumn{1}{c|}{0.0}          & \multicolumn{1}{c|}{0.92}                 & \multicolumn{1}{c|}{0.876}                      & \multicolumn{1}{c|}{0.952}                                  & 1.0                                          & 1.952                                                                                                        & \multicolumn{1}{c|}{0.96}                                  & \multicolumn{1}{c|}{1.0} & 0.88                                \\ \hline
Sales Lead Generation                 & \multicolumn{1}{c|}{0.015}          & \multicolumn{1}{c|}{0.039}          & \multicolumn{1}{c|}{0.92}                 & \multicolumn{1}{c|}{0.841}                      & \multicolumn{1}{c|}{0.887}                                  & 1.0                                          & 1.55                                                                                                        & \multicolumn{1}{c|}{-}                                & \multicolumn{1}{c|}{1.0} & 0.50                                \\ \hline
Customer Complaint                    & \multicolumn{1}{c|}{0.008}          & \multicolumn{1}{c|}{0.029}          & \multicolumn{1}{c|}{0.964}                 & \multicolumn{1}{c|}{1.0}                      & \multicolumn{1}{c|}{0.913}                                  & 1.0                                          & 1.36                                                                                                        & \multicolumn{1}{c|}{1.0}                                  & \multicolumn{1}{c|}{1.0} & 0.873                                \\ \hline
Passport Application                  & \multicolumn{1}{c|}{0.009}          & \multicolumn{1}{c|}{0.038}          & \multicolumn{1}{c|}{0.928}                 & \multicolumn{1}{c|}{0.986}                      & \multicolumn{1}{c|}{0.963}                                  & 1.0                                          & 1.604                                                                                                        & \multicolumn{1}{c|}{0.96}                                  & \multicolumn{1}{c|}{1.0} & 0.86                                \\ \hline
\textbf{Average}                      & \multicolumn{1}{c|}{\textbf{0.009}} & \multicolumn{1}{c|}{\textbf{0.038}} & \multicolumn{1}{c|}{\textbf{0.94}}       & \multicolumn{1}{c|}{\textbf{0.931}}                      & \multicolumn{1}{c|}{\textbf{0.933}}                         & \textbf{1.0}                                 & \textbf{1.433}                                                                                               & \multicolumn{1}{c|}{\textbf{0.985}}                        & \multicolumn{1}{c|}{\textbf{1.0}} & \textbf{0.75}                       \\ \hline
\end{tabular}%
}
\end{table*}

\section{Results and Discussions}
\label{sec:exp-results-discussions}

\textbf{Evaluation Metric}: To evaluate SmartFlow's accuracy in generating navigation workflows and entering correct values into data fields, we calculate the following metrics:
\begin{itemize}
    \item \textit{Text-extraction Accuracy}: Measures the accuracy of detecting text fields on the application screen using OCR techniques such as EasyOCR in terms of Character Error Rate (CER) and Word Error Rate (WER).
    \item \textit{Layout Mapping Accuracy}: Evaluates the correct association of field names with edit fields, placeholders, and data hints.
    \item \textit{Filled Data Accuracy}: Determines the accuracy of filling fields in the application form with correct data values.
    \item \textit{Request Submission Accuracy}: This metric measures the success or failure of executing the task request.
    \item \textit{Complex Component Accuracy}: Reports the accuracy of filling data in complex fields such as datepickers, dropdowns, radio buttons and checkboxes.
    \item \textit{Task Completion Time}: This measures the time (in minutes) taken to complete one specific task-request.
    
\end{itemize}

\noindent \textbf{Experimental Results}: We conducted our experiments using OpenAI's LLM GPT-3~\cite{gpt3} API, which is publicly available. The experiments were performed on a GTX 1080 machine with 8 GB GPU Memory. In Table~\ref{tab:results-cas}, we present the performance results of SmartFlow on the Conference Attendance System (CAS), a two-page web application with diverse layouts. The text extraction accuracy of OCR is high, with an average CER of $0.015$ and WER of $0.086$. We also compare the accuracy of the rule-based and virtual-grid layout mapping approaches, which show similar and satisfactory results. The minor mistakes in layout mapping can be attributed to certain factors such as closeness of field name and/or hint with incorrect edit-field, cascaded OCR text detection error. These errors were corrected during the initial setup of SmartFlow on the system by Admin. The accuracy of filled data is $95\%$, with errors primarily occurring in radio-button and checkbox fields. Finetuning LLMs to handle these fields would significantly improve the accuracy of filled data. The request submission accuracy is $100\%$, indicating that SmartFlow accurately reads the status of executed requests. The average task completion time for CAS is $5.7$ minutes, considering its multi-page nature. Variations in task completion time across different layouts and user-tasks are mainly influenced by datepicker selections and scrolling within dropdown fields. We also report the results for Patient Registration application in Table~\ref{tab:results-patient}. Please refer supplementary material~\footnote{Supplementary File: \url{https://smartflow-4c5a0a.webflow.io/}} for additional results.

Next, we report the average accuracy for each application of the RPA-dataset with different layouts in Table~\ref{tab:results-apps}. 
In Table~\ref{tab:results-apps}, it is evident that SmartFlow efficiently automated various applications with an average filled data accuracy of $93.3\%$ and an average time to submit requests of $1.433$ minutes. The main challenge lies in accurately selecting options for radio-buttons and checkboxes which can be achieved by fine-tuning the LLMs with such data fields.

\section{Limitations of SmartFlow}
\label{sec:limitations}
In this section, we mention the limitations of SmartFlow:
\vspace{-1mm}
\begin{itemize}
\item {Dynamic fields}: In the current version, handling dynamic fields is not supported. One approach is to generate PyAutoGui code for each field individually and perform layout mapping after filling data in that field. However, this method is time-consuming and inefficient. We are actively researching more efficient solutions for dynamic field handling.

\item \textit{Scrollable forms}: We can use the HTML source code to determine if the page is scrollable and then capture screenshots of the visible fields, perform layout mapping and generate PyAutoGui code for filling them. If there are hidden fields or buttons such as "next" or "submit," we scroll down the page and repeat the process until we locate the submit button.

\item \textit{Inference of field types}: Future versions of SmartFlow aim to enhance the accuracy and flexibility of determining field types by training a deep learning-based object detector. This will reduce reliance on the HTML source code alone. However, due to the unavailability of training data, this feature is not included in the current version.
\end{itemize}

\vspace{-3mm}
\section{Conclusion and Future Work}
\label{sec:conclusion}
This paper introduced SmartFlow, an AI-driven training-free RPA system designed to autonomously execute user task-requests. By integrating computer vision and generative models such as LLMs, SmartFlow is able to automatically generate navigation workflows and adapt to variations in GUIs and applications without human intervention. Our experiments on a self-created RPA-dataset, consisting of diverse web applications with varying layouts and user task-requests, showcased the impressive performance of SmartFlow. Moving forward, our future work will focus on handling dynamic web-applications with scrollable forms and training a deep-learning-based object detector to infer field types, reducing the reliance on HTML code.

\bibliographystyle{ACM-Reference-Format}
\bibliography{main.bib}


\begin{thebibliography}{34}


\ifx \showCODEN    \undefined \def \showCODEN     #1{\unskip}     \fi
\ifx \showDOI      \undefined \def \showDOI       #1{#1}\fi
\ifx \showISBNx    \undefined \def \showISBNx     #1{\unskip}     \fi
\ifx \showISBNxiii \undefined \def \showISBNxiii  #1{\unskip}     \fi
\ifx \showISSN     \undefined \def \showISSN      #1{\unskip}     \fi
\ifx \showLCCN     \undefined \def \showLCCN      #1{\unskip}     \fi
\ifx \shownote     \undefined \def \shownote      #1{#1}          \fi
\ifx \showarticletitle \undefined \def \showarticletitle #1{#1}   \fi
\ifx \showURL      \undefined \def \showURL       {\relax}        \fi
\providecommand\bibfield[2]{#2}
\providecommand\bibinfo[2]{#2}
\providecommand\natexlab[1]{#1}
\providecommand\showeprint[2][]{arXiv:#2}

\bibitem[Aguirre and Rodr{\'i}guez(2017)]%
        {rpa-service1}
\bibfield{author}{\bibinfo{person}{Santiago Aguirre} {and} \bibinfo{person}{Alejandro Rodr{\'i}guez}.} \bibinfo{year}{2017}\natexlab{}.
\newblock \showarticletitle{Automation of a Business Process Using Robotic Process Automation (RPA): A Case Study}. In \bibinfo{booktitle}{\emph{Workshop on Engineering Applications}}.
\newblock


\bibitem[Anagnoste(2018)]%
        {rpa-hr2}
\bibfield{author}{\bibinfo{person}{Sorin Anagnoste}.} \bibinfo{year}{2018}\natexlab{}.
\newblock \showarticletitle{Robotic Automation Process – The operating system for the digital enterprise}.
\newblock \bibinfo{journal}{\emph{Proceedings of the International Conference on Business Excellence}}  \bibinfo{volume}{12} (\bibinfo{year}{2018}), \bibinfo{pages}{54 -- 69}.
\newblock


\bibitem[Anywhere(2003)]%
        {automation-anywhere}
\bibfield{author}{\bibinfo{person}{Automation Anywhere}.} \bibinfo{year}{2003}\natexlab{}.
\newblock \bibinfo{title}{The automation success platform}.
\newblock \bibinfo{howpublished}{\url{https://www.automationanywhere.com/}}.
\newblock
\newblock
\shownote{Accessed: 2023-05-15}.


\bibitem[Brown et~al\mbox{.}(2020)]%
        {gpt3}
\bibfield{author}{\bibinfo{person}{Tom~B. Brown}, \bibinfo{person}{Benjamin Mann}, \bibinfo{person}{Nick Ryder}, \bibinfo{person}{Melanie Subbiah}, \bibinfo{person}{Jared Kaplan}, \bibinfo{person}{Prafulla Dhariwal}, \bibinfo{person}{Arvind Neelakantan}, \bibinfo{person}{Pranav Shyam}, \bibinfo{person}{Girish Sastry}, \bibinfo{person}{Amanda Askell}, \bibinfo{person}{Sandhini Agarwal}, \bibinfo{person}{Ariel Herbert-Voss}, \bibinfo{person}{Gretchen Krueger}, \bibinfo{person}{Tom Henighan}, \bibinfo{person}{Rewon Child}, \bibinfo{person}{Aditya Ramesh}, \bibinfo{person}{Daniel~M. Ziegler}, \bibinfo{person}{Jeffrey Wu}, \bibinfo{person}{Clemens Winter}, \bibinfo{person}{Christopher Hesse}, \bibinfo{person}{Mark Chen}, \bibinfo{person}{Eric Sigler}, \bibinfo{person}{Mateusz Litwin}, \bibinfo{person}{Scott Gray}, \bibinfo{person}{Benjamin Chess}, \bibinfo{person}{Jack Clark}, \bibinfo{person}{Christopher Berner}, \bibinfo{person}{Sam McCandlish}, \bibinfo{person}{Alec Radford}, \bibinfo{person}{Ilya Sutskever},
  {and} \bibinfo{person}{Dario Amodei}.} \bibinfo{year}{2020}\natexlab{}.
\newblock \bibinfo{title}{Language Models are Few-Shot Learners}.
\newblock
\newblock
\showeprint[arxiv]{2005.14165}~[cs.CL]


\bibitem[Brown et~al\mbox{.}(2022)]%
        {chatgpt}
\bibfield{author}{\bibinfo{person}{Tom~B. Brown}, \bibinfo{person}{Benjamin Mann}, \bibinfo{person}{Nick Ryder}, \bibinfo{person}{Muthukumar Subbiah}, \bibinfo{person}{Jared Kaplan}, \bibinfo{person}{Prafulla Dhariwal}, \bibinfo{person}{Kalyan Neelakantan}, \bibinfo{person}{Dario Amodei}, {and} \bibinfo{person}{Ilya Sutskever}.} \bibinfo{year}{2022}\natexlab{}.
\newblock \showarticletitle{ChatGPT: Optimizing Language Models for Dialogue}.
\newblock \bibinfo{journal}{\emph{arXiv preprint arXiv:2201.08237}} (\bibinfo{year}{2022}).
\newblock


\bibitem[Carion et~al\mbox{.}(2020)]%
        {detr}
\bibfield{author}{\bibinfo{person}{Nicolas Carion}, \bibinfo{person}{Francisco Massa}, \bibinfo{person}{Gabriel Synnaeve}, \bibinfo{person}{Nicolas Usunier}, \bibinfo{person}{Alexander Kirillov}, {and} \bibinfo{person}{Sergey Zagoruyko}.} \bibinfo{year}{2020}\natexlab{}.
\newblock \showarticletitle{End-to-End Object Detection with Transformers}.
\newblock \bibinfo{journal}{\emph{CoRR}}  \bibinfo{volume}{abs/2005.12872} (\bibinfo{year}{2020}).
\newblock
\showeprint[arXiv]{2005.12872}
\urldef\tempurl%
\url{https://arxiv.org/abs/2005.12872}
\showURL{%
\tempurl}


\bibitem[Chowdhery et~al\mbox{.}(2022)]%
        {palm}
\bibfield{author}{\bibinfo{person}{Aakanksha Chowdhery}, \bibinfo{person}{Sharan Narang}, \bibinfo{person}{Jacob Devlin}, \bibinfo{person}{Maarten Bosma}, \bibinfo{person}{Gaurav Mishra}, \bibinfo{person}{Adam Roberts}, \bibinfo{person}{Paul Barham}, \bibinfo{person}{Hyung~Won Chung}, \bibinfo{person}{Charles Sutton}, \bibinfo{person}{Sebastian Gehrmann}, \bibinfo{person}{Parker Schuh}, \bibinfo{person}{Kensen Shi}, \bibinfo{person}{Sasha Tsvyashchenko}, \bibinfo{person}{Joshua Maynez}, \bibinfo{person}{Abhishek Rao}, \bibinfo{person}{Parker Barnes}, \bibinfo{person}{Yi Tay}, \bibinfo{person}{Noam Shazeer}, \bibinfo{person}{Vinodkumar Prabhakaran}, \bibinfo{person}{Emily Reif}, \bibinfo{person}{Nan Du}, \bibinfo{person}{Ben Hutchinson}, \bibinfo{person}{Reiner Pope}, \bibinfo{person}{James Bradbury}, \bibinfo{person}{Jacob Austin}, \bibinfo{person}{Michael Isard}, \bibinfo{person}{Guy Gur-Ari}, \bibinfo{person}{Pengcheng Yin}, \bibinfo{person}{Toju Duke}, \bibinfo{person}{Anselm Levskaya},
  \bibinfo{person}{Sanjay Ghemawat}, \bibinfo{person}{Sunipa Dev}, \bibinfo{person}{Henryk Michalewski}, \bibinfo{person}{Xavier Garcia}, \bibinfo{person}{Vedant Misra}, \bibinfo{person}{Kevin Robinson}, \bibinfo{person}{Liam Fedus}, \bibinfo{person}{Denny Zhou}, \bibinfo{person}{Daphne Ippolito}, \bibinfo{person}{David Luan}, \bibinfo{person}{Hyeontaek Lim}, \bibinfo{person}{Barret Zoph}, \bibinfo{person}{Alexander Spiridonov}, \bibinfo{person}{Ryan Sepassi}, \bibinfo{person}{David Dohan}, \bibinfo{person}{Shivani Agrawal}, \bibinfo{person}{Mark Omernick}, \bibinfo{person}{Andrew~M. Dai}, \bibinfo{person}{Thanumalayan~Sankaranarayana Pillai}, \bibinfo{person}{Marie Pellat}, \bibinfo{person}{Aitor Lewkowycz}, \bibinfo{person}{Erica Moreira}, \bibinfo{person}{Rewon Child}, \bibinfo{person}{Oleksandr Polozov}, \bibinfo{person}{Katherine Lee}, \bibinfo{person}{Zongwei Zhou}, \bibinfo{person}{Xuezhi Wang}, \bibinfo{person}{Brennan Saeta}, \bibinfo{person}{Mark Diaz}, \bibinfo{person}{Orhan Firat},
  \bibinfo{person}{Michele Catasta}, \bibinfo{person}{Jason Wei}, \bibinfo{person}{Kathy Meier-Hellstern}, \bibinfo{person}{Douglas Eck}, \bibinfo{person}{Jeff Dean}, \bibinfo{person}{Slav Petrov}, {and} \bibinfo{person}{Noah Fiedel}.} \bibinfo{year}{2022}\natexlab{}.
\newblock \bibinfo{title}{PaLM: Scaling Language Modeling with Pathways}.
\newblock
\newblock
\showeprint[arxiv]{2204.02311}~[cs.CL]


\bibitem[D et~al\mbox{.}(2018)]%
        {deepreader}
\bibfield{author}{\bibinfo{person}{Vishwanath D}, \bibinfo{person}{Rohit Rahul}, \bibinfo{person}{Gunjan Sehgal}, \bibinfo{person}{Swati}, \bibinfo{person}{Arindam Chowdhury}, \bibinfo{person}{Monika Sharma}, \bibinfo{person}{Lovekesh Vig}, \bibinfo{person}{Gautam Shroff}, {and} \bibinfo{person}{Ashwin Srinivasan}.} \bibinfo{year}{2018}\natexlab{}.
\newblock \bibinfo{title}{Deep Reader: Information extraction from Document images via relation extraction and Natural Language}.
\newblock
\newblock
\showeprint[arxiv]{1812.04377}~[cs.CV]


\bibitem[Flechsig et~al\mbox{.}(2022)]%
        {rpa-supply}
\bibfield{author}{\bibinfo{person}{Christian Flechsig}, \bibinfo{person}{Franziska Anslinger}, {and} \bibinfo{person}{Rainer Lasch}.} \bibinfo{year}{2022}\natexlab{}.
\newblock \showarticletitle{Robotic Process Automation in purchasing and supply management: A multiple case study on potentials, barriers, and implementation}.
\newblock \bibinfo{journal}{\emph{Journal of Purchasing and Supply Management}} \bibinfo{volume}{28}, \bibinfo{number}{1} (\bibinfo{year}{2022}), \bibinfo{pages}{100718}.
\newblock
\showISSN{1478-4092}
\urldef\tempurl%
\url{https://doi.org/10.1016/j.pursup.2021.100718}
\showDOI{\tempurl}


\bibitem[Geyer‐Klingeberg et~al\mbox{.}(2018)]%
        {rpa4}
\bibfield{author}{\bibinfo{person}{Jerome Geyer‐Klingeberg}, \bibinfo{person}{Janina Nakladal}, \bibinfo{person}{Fabian Baldauf}, {and} \bibinfo{person}{Fabian Veit}.} \bibinfo{year}{2018}\natexlab{}.
\newblock \showarticletitle{Process Mining and Robotic Process Automation: A Perfect Match}. In \bibinfo{booktitle}{\emph{International Conference on Business Process Management}}.
\newblock


\bibitem[Gupta et~al\mbox{.}(2018)]%
        {rpa-hr1}
\bibfield{author}{\bibinfo{person}{Pooja Gupta}, \bibinfo{person}{Semila Fernandes}, {and} \bibinfo{person}{Manish Jain}.} \bibinfo{year}{2018}\natexlab{}.
\newblock \showarticletitle{Automation in recruitment: a new frontier}.
\newblock \bibinfo{journal}{\emph{Journal of Information Technology Teaching Cases}}  \bibinfo{volume}{8} (\bibinfo{year}{2018}), \bibinfo{pages}{118--125}.
\newblock


\bibitem[Hallikainen et~al\mbox{.}(2018)]%
        {rpa-service2}
\bibfield{author}{\bibinfo{person}{Petri Hallikainen}, \bibinfo{person}{Riitta Bekkhus}, {and} \bibinfo{person}{Shan Pan}.} \bibinfo{year}{2018}\natexlab{}.
\newblock \showarticletitle{How OpusCapita Used Internal RPA Capabilities to Offer Services to Clients}.
\newblock \bibinfo{journal}{\emph{MIS Q. Executive}}  \bibinfo{volume}{17} (\bibinfo{year}{2018}).
\newblock


\bibitem[Ivan{\v{c}}i{\'{c}} et~al\mbox{.}(2019)]%
        {rpa-review}
\bibfield{author}{\bibinfo{person}{Lucija Ivan{\v{c}}i{\'{c}}}, \bibinfo{person}{Dalia Su{\v{s}}a~Vugec}, {and} \bibinfo{person}{Vesna Bosilj~Vuk{\v{s}}i{\'{c}}}.} \bibinfo{year}{2019}\natexlab{}.
\newblock \showarticletitle{Robotic Process Automation: Systematic Literature Review}. In \bibinfo{booktitle}{\emph{International Conference on Business Process Management}}.
\newblock


\bibitem[Kirillov et~al\mbox{.}(2023)]%
        {segment-anything}
\bibfield{author}{\bibinfo{person}{Alexander Kirillov}, \bibinfo{person}{Eric Mintun}, \bibinfo{person}{Nikhila Ravi}, \bibinfo{person}{Hanzi Mao}, \bibinfo{person}{Chloe Rolland}, \bibinfo{person}{Laura Gustafson}, \bibinfo{person}{Tete Xiao}, \bibinfo{person}{Spencer Whitehead}, \bibinfo{person}{Alexander~C. Berg}, \bibinfo{person}{Wan-Yen Lo}, \bibinfo{person}{Piotr Dollár}, {and} \bibinfo{person}{Ross Girshick}.} \bibinfo{year}{2023}\natexlab{}.
\newblock \bibinfo{title}{Segment Anything}.
\newblock
\newblock
\showeprint[arxiv]{2304.02643}~[cs.CV]


\bibitem[Leshob et~al\mbox{.}(2018)]%
        {rpa3}
\bibfield{author}{\bibinfo{person}{Abderrahmane Leshob}, \bibinfo{person}{Audrey Bourgouin}, {and} \bibinfo{person}{Laurent Renard}.} \bibinfo{year}{2018}\natexlab{}.
\newblock \showarticletitle{Towards a Process Analysis Approach to Adopt Robotic Process Automation}.
\newblock \bibinfo{journal}{\emph{2018 IEEE 15th International Conference on e-Business Engineering (ICEBE)}} (\bibinfo{year}{2018}), \bibinfo{pages}{46--53}.
\newblock


\bibitem[Liang(2021)]%
        {video-analysis}
\bibfield{author}{\bibinfo{person}{Junwei Liang}.} \bibinfo{year}{2021}\natexlab{}.
\newblock \bibinfo{title}{From Recognition to Prediction: Analysis of Human Action and Trajectory Prediction in Video}.
\newblock
\newblock
\showeprint[arxiv]{2011.10670}~[cs.CV]


\bibitem[Madakam et~al\mbox{.}(2019)]%
        {rpa2}
\bibfield{author}{\bibinfo{person}{Somayya Madakam}, \bibinfo{person}{Rajesh~M. Holmukhe}, {and} \bibinfo{person}{Durgesh~Kumar Jaiswal}.} \bibinfo{year}{2019}\natexlab{}.
\newblock \showarticletitle{The Future Digital Work Force: Robotic Process Automation (RPA)}.
\newblock \bibinfo{journal}{\emph{Journal of Information Systems and Technology Management}} (\bibinfo{year}{2019}).
\newblock


\bibitem[Martins et~al\mbox{.}(2020)]%
        {rpa-ml}
\bibfield{author}{\bibinfo{person}{Pedro Martins}, \bibinfo{person}{Filipe Sá}, \bibinfo{person}{Francisco Morgado}, {and} \bibinfo{person}{Carlos Cunha}.} \bibinfo{year}{2020}\natexlab{}.
\newblock \showarticletitle{Using machine learning for cognitive Robotic Process Automation (RPA)}. In \bibinfo{booktitle}{\emph{2020 15th Iberian Conference on Information Systems and Technologies (CISTI)}}. \bibinfo{pages}{1--6}.
\newblock
\urldef\tempurl%
\url{https://doi.org/10.23919/CISTI49556.2020.9140440}
\showDOI{\tempurl}


\bibitem[OpenAI(2023)]%
        {gpt4}
\bibfield{author}{\bibinfo{person}{OpenAI}.} \bibinfo{year}{2023}\natexlab{}.
\newblock \bibinfo{title}{GPT-4 Technical Report}.
\newblock
\newblock
\showeprint[arxiv]{2303.08774}~[cs.CL]


\bibitem[Palm et~al\mbox{.}(2021)]%
        {attend}
\bibfield{author}{\bibinfo{person}{Rasmus~Berg Palm}, \bibinfo{person}{Florian Laws}, {and} \bibinfo{person}{Ole Winther}.} \bibinfo{year}{2021}\natexlab{}.
\newblock \bibinfo{title}{Attend, Copy, Parse -- End-to-end information extraction from documents}.
\newblock
\newblock
\showeprint[arxiv]{1812.07248}~[cs.CL]


\bibitem[Penttinen et~al\mbox{.}(2018)]%
        {rpa5}
\bibfield{author}{\bibinfo{person}{Esko Penttinen}, \bibinfo{person}{Henje Kasslin}, {and} \bibinfo{person}{Aleksandre Asatiani}.} \bibinfo{year}{2018}\natexlab{}.
\newblock \showarticletitle{How to Choose between Robotic Process Automation and Back-end System Automation?}. In \bibinfo{booktitle}{\emph{European Conference on Information Systems}}.
\newblock


\bibitem[Prism(2001)]%
        {blueprism}
\bibfield{author}{\bibinfo{person}{SS\&C~Blue Prism}.} \bibinfo{year}{2001}\natexlab{}.
\newblock \bibinfo{title}{Intelligent Robotic Process Automation}.
\newblock \bibinfo{howpublished}{\url{https://www.blueprism.com/}}.
\newblock
\newblock
\shownote{Accessed: 2023-05-15}.


\bibitem[Ratia et~al\mbox{.}(2018)]%
        {rpa-health1}
\bibfield{author}{\bibinfo{person}{M. Ratia}, \bibinfo{person}{Jussi Myll{\"a}rniemi}, {and} \bibinfo{person}{Nina Helander}.} \bibinfo{year}{2018}\natexlab{}.
\newblock \showarticletitle{Robotic Process Automation - Creating Value by Digitalizing Work in the Private Healthcare?}
\newblock \bibinfo{journal}{\emph{Proceedings of the 22nd International Academic Mindtrek Conference}} (\bibinfo{year}{2018}).
\newblock


\bibitem[Redmon et~al\mbox{.}(2016)]%
        {yolo}
\bibfield{author}{\bibinfo{person}{Joseph Redmon}, \bibinfo{person}{Santosh Divvala}, \bibinfo{person}{Ross Girshick}, {and} \bibinfo{person}{Ali Farhadi}.} \bibinfo{year}{2016}\natexlab{}.
\newblock \bibinfo{title}{You Only Look Once: Unified, Real-Time Object Detection}.
\newblock
\newblock
\showeprint[arxiv]{1506.02640}~[cs.CV]


\bibitem[Romao et~al\mbox{.}(2019)]%
        {rpa-banking1}
\bibfield{author}{\bibinfo{person}{Mário Romao}, \bibinfo{person}{Joao Costa}, {and} \bibinfo{person}{Carlos~J. Costa}.} \bibinfo{year}{2019}\natexlab{}.
\newblock \showarticletitle{Robotic Process Automation: A Case Study in the Banking Industry}. In \bibinfo{booktitle}{\emph{2019 14th Iberian Conference on Information Systems and Technologies (CISTI)}}. \bibinfo{pages}{1--6}.
\newblock
\urldef\tempurl%
\url{https://doi.org/10.23919/CISTI.2019.8760733}
\showDOI{\tempurl}


\bibitem[Sharma(2014)]%
        {selenium}
\bibfield{author}{\bibinfo{person}{Monika Sharma}.} \bibinfo{year}{2014}\natexlab{}.
\newblock \showarticletitle{Selenium Tool: A Web based Automation Testing Framework}.
\newblock


\bibitem[Siderska(2020)]%
        {rpa1}
\bibfield{author}{\bibinfo{person}{Julia Siderska}.} \bibinfo{year}{2020}\natexlab{}.
\newblock \showarticletitle{Robotic Process Automation — a driver of digital transformation?}
\newblock \bibinfo{journal}{\emph{Engineering Management in Production and Services}}  \bibinfo{volume}{12} (\bibinfo{year}{2020}), \bibinfo{pages}{21 -- 31}.
\newblock


\bibitem[Touvron et~al\mbox{.}(2023)]%
        {llama}
\bibfield{author}{\bibinfo{person}{Hugo Touvron}, \bibinfo{person}{Thibaut Lavril}, \bibinfo{person}{Gautier Izacard}, \bibinfo{person}{Xavier Martinet}, \bibinfo{person}{Marie-Anne Lachaux}, \bibinfo{person}{Timoth{\'e}e Lacroix}, \bibinfo{person}{Baptiste Rozi{\`e}re}, \bibinfo{person}{Naman Goyal}, \bibinfo{person}{Eric Hambro}, \bibinfo{person}{Faisal Azhar}, \bibinfo{person}{Aurelien Rodriguez}, \bibinfo{person}{Armand Joulin}, \bibinfo{person}{Edouard Grave}, {and} \bibinfo{person}{Guillaume Lample}.} \bibinfo{year}{2023}\natexlab{}.
\newblock \showarticletitle{LLaMA: Open and Efficient Foundation Language Models}.
\newblock \bibinfo{journal}{\emph{arXiv preprint arXiv:2302.13971}} (\bibinfo{year}{2023}).
\newblock


\bibitem[UiPath(2005)]%
        {uipath}
\bibfield{author}{\bibinfo{person}{UiPath}.} \bibinfo{year}{2005}\natexlab{}.
\newblock \bibinfo{title}{AI-powered business automation}.
\newblock \bibinfo{howpublished}{\url{https://www.uipath.com/}}.
\newblock
\newblock
\shownote{Accessed: 2023-05-15}.


\bibitem[Wang et~al\mbox{.}(2023)]%
        {mobileui-llm}
\bibfield{author}{\bibinfo{person}{Bryan Wang}, \bibinfo{person}{Gang Li}, {and} \bibinfo{person}{Yang Li}.} \bibinfo{year}{2023}\natexlab{}.
\newblock \bibinfo{title}{Enabling Conversational Interaction with Mobile UI using Large Language Models}.
\newblock
\newblock
\showeprint[arxiv]{2209.08655}~[cs.HC]


\bibitem[Willcocks et~al\mbox{.}(2017)]%
        {rpa-banking2}
\bibfield{author}{\bibinfo{person}{Leslie~P. Willcocks}, \bibinfo{person}{Mary~C Lacity}, {and} \bibinfo{person}{Andrew Craig}.} \bibinfo{year}{2017}\natexlab{}.
\newblock \showarticletitle{Robotic process automation: strategic transformation lever for global business services?}
\newblock \bibinfo{journal}{\emph{Journal of Information Technology Teaching Cases}}  \bibinfo{volume}{7} (\bibinfo{year}{2017}), \bibinfo{pages}{17--28}.
\newblock


\bibitem[Wu et~al\mbox{.}(2023)]%
        {visual-chatgpt}
\bibfield{author}{\bibinfo{person}{Chenfei Wu}, \bibinfo{person}{Shengming Yin}, \bibinfo{person}{Weizhen Qi}, \bibinfo{person}{Xiaodong Wang}, \bibinfo{person}{Zecheng Tang}, {and} \bibinfo{person}{Nan Duan}.} \bibinfo{year}{2023}\natexlab{}.
\newblock \bibinfo{title}{Visual ChatGPT: Talking, Drawing and Editing with Visual Foundation Models}.
\newblock
\newblock
\showeprint[arxiv]{2303.04671}~[cs.CV]


\bibitem[Yu et~al\mbox{.}(2020)]%
        {pick}
\bibfield{author}{\bibinfo{person}{Wenwen Yu}, \bibinfo{person}{Ning Lu}, \bibinfo{person}{Xianbiao Qi}, \bibinfo{person}{Ping Gong}, {and} \bibinfo{person}{Rong Xiao}.} \bibinfo{year}{2020}\natexlab{}.
\newblock \bibinfo{title}{PICK: Processing Key Information Extraction from Documents using Improved Graph Learning-Convolutional Networks}.
\newblock
\newblock
\showeprint[arxiv]{2004.07464}~[cs.CV]


\bibitem[Zhang and Agrawala(2023)]%
        {control-net}
\bibfield{author}{\bibinfo{person}{Lvmin Zhang} {and} \bibinfo{person}{Maneesh Agrawala}.} \bibinfo{year}{2023}\natexlab{}.
\newblock \bibinfo{title}{Adding Conditional Control to Text-to-Image Diffusion Models}.
\newblock
\newblock
\showeprint[arxiv]{2302.05543}~[cs.CV]


\end{thebibliography}

\appendix
\section{Supplementary}

\begin{figure*}[h!]
  \centering
  \includegraphics[width=0.9\textwidth]{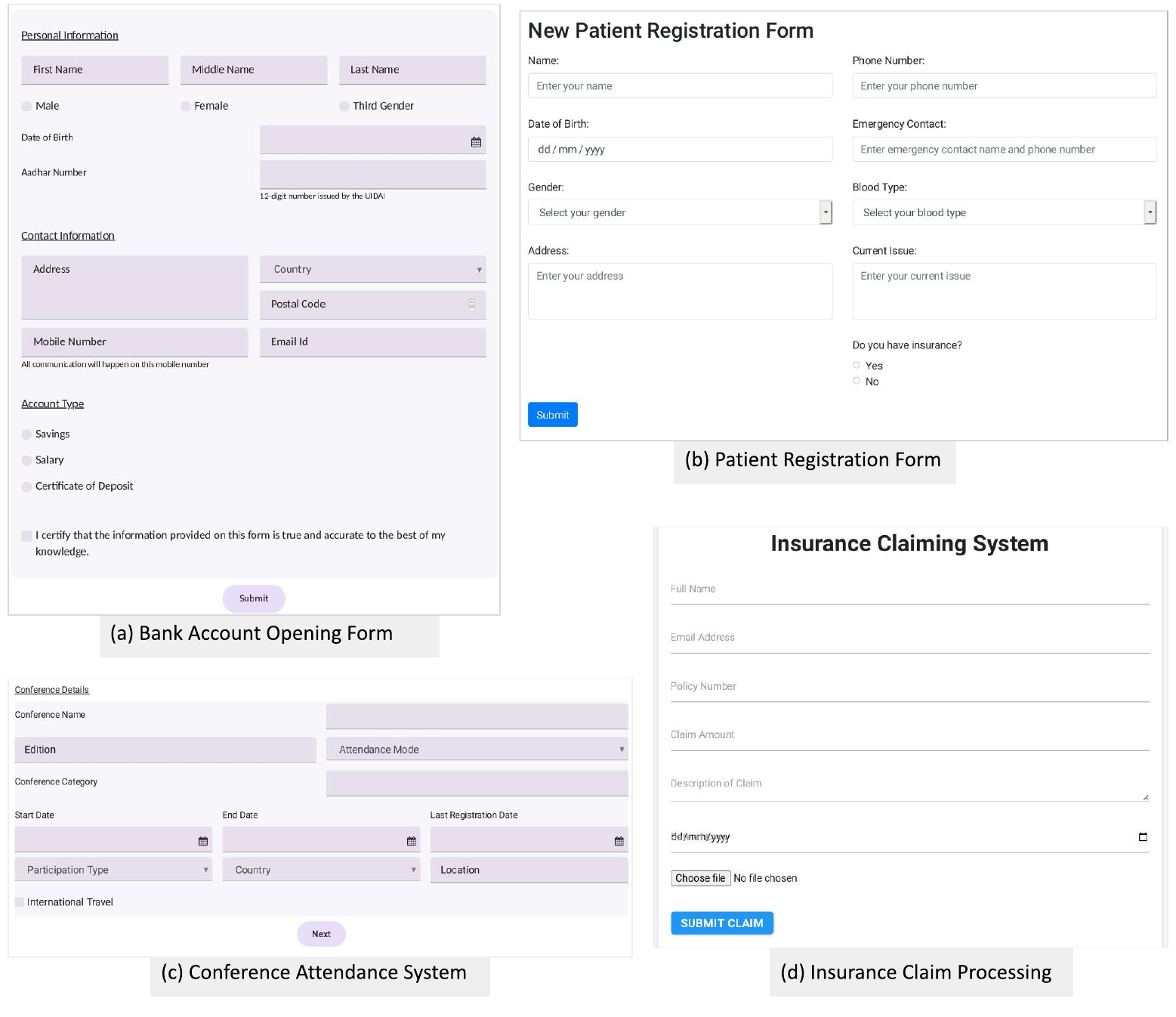}
  \caption{Figure showing examples of diverse application layouts.}
  \label{fig:app-layouts}
\end{figure*}

\begin{table*}[]
\centering
\caption{Table showing the performance of SmartFlow on Sales Lead Generation application.}
\label{tab:results-sales}
\resizebox{\textwidth}{!}{%
\begin{tabular}{|c|c|cccccc|c|ccc|}
\hline
\multirow{3}{*}{\textbf{Layout No.}} &
  \multirow{3}{*}{\textbf{Page No.}} &
  \multicolumn{6}{c|}{\textbf{Accuracy}} &
  \multirow{3}{*}{\textbf{\begin{tabular}[c]{@{}c@{}}Task completion average time\\ (in mins)\end{tabular}}} &
  \multicolumn{3}{c|}{\textbf{Complex Component Accuracy}} \\ \cline{3-8} \cline{10-12} 
 &
   &
  \multicolumn{2}{c|}{\textbf{OCR}} &
  \multicolumn{2}{c|}{\textbf{Layout Mapping}} &
  \multicolumn{1}{c|}{\multirow{2}{*}{\textbf{Filled Data}}} &
  \multirow{2}{*}{\textbf{Request Submission}} &
   &
  \multicolumn{1}{c|}{\multirow{2}{*}{\textbf{Datepicker}}} &
  \multicolumn{1}{c|}{\multirow{2}{*}{\textbf{Dropdown}}} &
  \multirow{2}{*}{\textbf{Radio/Checkbox}} \\ \cline{3-6}
 &
   &
  \multicolumn{1}{c|}{\textbf{CER}} &
  \multicolumn{1}{c|}{\textbf{WER}} &
  \multicolumn{1}{c|}{\textbf{Rule-based}} &
  \multicolumn{1}{c|}{\textbf{Virtual-Grid}} &
  \multicolumn{1}{c|}{} &
   &
   &
  \multicolumn{1}{c|}{} &
  \multicolumn{1}{c|}{} &
   \\ \hline
1 &
  1 &
  \multicolumn{1}{c|}{0.044} &
  \multicolumn{1}{c|}{0.083} &
  \multicolumn{1}{c|}{0.727} &
  \multicolumn{1}{c|}{0.636} &
  \multicolumn{1}{c|}{0.885} &
  1.0 &
  1.37 &
  \multicolumn{1}{c|}{-} &
  \multicolumn{1}{c|}{1.0} &
  0.6 \\ \hline
2 &
  1 &
  \multicolumn{1}{c|}{0.0} &
  \multicolumn{1}{c|}{0.0} &
  \multicolumn{1}{c|}{1.0} &
  \multicolumn{1}{c|}{0.833} &
  \multicolumn{1}{c|}{0.885} &
  1.0 &
  1.38 &
  \multicolumn{1}{c|}{-} &
  \multicolumn{1}{c|}{1.0} &
  0.5 \\ \hline
3 &
  1 &
  \multicolumn{1}{c|}{0.003} &
  \multicolumn{1}{c|}{0.016} &
  \multicolumn{1}{c|}{0.875} &
  \multicolumn{1}{c|}{1.0} &
  \multicolumn{1}{c|}{0.925} &
  1.0 &
  1.71 &
  \multicolumn{1}{c|}{-} &
  \multicolumn{1}{c|}{1.0} &
  0.7 \\ \hline
4 &
  1 &
  \multicolumn{1}{c|}{0.0} &
  \multicolumn{1}{c|}{0.0} &
  \multicolumn{1}{c|}{1.0} &
  \multicolumn{1}{c|}{1.0} &
  \multicolumn{1}{c|}{0.9} &
  1.0 &
  1.12 &
  \multicolumn{1}{c|}{-} &
  \multicolumn{1}{c|}{1.0} &
  0.4 \\ \hline
5 &
  1 &
  \multicolumn{1}{c|}{0.031} &
  \multicolumn{1}{c|}{0.099} &
  \multicolumn{1}{c|}{1.0} &
  \multicolumn{1}{c|}{0.733} &
  \multicolumn{1}{c|}{0.844} &
  1.0 &
  2.2 &
  \multicolumn{1}{c|}{-} &
  \multicolumn{1}{c|}{1.0} &
  0.3 \\ \hline
\textbf{Average} &
  \textbf{} &
  \multicolumn{1}{c|}{\textbf{0.015}} &
  \multicolumn{1}{c|}{\textbf{0.039}} &
  \multicolumn{1}{c|}{\textbf{0.92}} &
  \multicolumn{1}{c|}{\textbf{0.841}} &
  \multicolumn{1}{c|}{\textbf{0.887}} &
  \textbf{1.0} &
  \textbf{1.55} &
  \multicolumn{1}{c|}{\textbf{-}} &
  \multicolumn{1}{c|}{\textbf{1.0}} &
  \textbf{0.5} \\ \hline
\end{tabular}%
}
\end{table*}

\begin{table*}[]
\centering
\caption{Table showing the performance of SmartFlow on Customer Complaint application.}
\label{tab:results-customer}
\resizebox{\textwidth}{!}{%
\begin{tabular}{|c|c|cccccc|c|ccc|}
\hline
\multirow{3}{*}{\textbf{Layout No.}} &
  \multirow{3}{*}{\textbf{Page No.}} &
  \multicolumn{6}{c|}{\textbf{Accuracy}} &
  \multirow{3}{*}{\textbf{\begin{tabular}[c]{@{}c@{}}Task completion average time\\ (in mins)\end{tabular}}} &
  \multicolumn{3}{c|}{\textbf{Complex Component Accuracy}} \\ \cline{3-8} \cline{10-12} 
 &
   &
  \multicolumn{2}{c|}{\textbf{OCR}} &
  \multicolumn{2}{c|}{\textbf{Layout Mapping}} &
  \multicolumn{1}{c|}{\multirow{2}{*}{\textbf{Filled Data}}} &
  \multirow{2}{*}{\textbf{Request Submission}} &
   &
  \multicolumn{1}{c|}{\multirow{2}{*}{\textbf{Datepicker}}} &
  \multicolumn{1}{c|}{\multirow{2}{*}{\textbf{Dropdown}}} &
  \multirow{2}{*}{\textbf{Radio/Checkbox}} \\ \cline{3-6}
 &
   &
  \multicolumn{1}{c|}{\textbf{CER}} &
  \multicolumn{1}{c|}{\textbf{WER}} &
  \multicolumn{1}{c|}{\textbf{Rule-based}} &
  \multicolumn{1}{c|}{\textbf{Virtual-Grid}} &
  \multicolumn{1}{c|}{} &
   &
   &
  \multicolumn{1}{c|}{} &
  \multicolumn{1}{c|}{} &
   \\ \hline
1 &
  1 &
  \multicolumn{1}{c|}{0.010} &
  \multicolumn{1}{c|}{0.033} &
  \multicolumn{1}{c|}{0.9} &
  \multicolumn{1}{c|}{1.0} &
  \multicolumn{1}{c|}{0.771} &
  1.0 &
  1.19 &
  \multicolumn{1}{c|}{-} &
  \multicolumn{1}{c|}{1.0} &
  0.5 \\ \hline
2 &
  1 &
  \multicolumn{1}{c|}{0.011} &
  \multicolumn{1}{c|}{0.035} &
  \multicolumn{1}{c|}{0.92} &
  \multicolumn{1}{c|}{1.0} &
  \multicolumn{1}{c|}{0.96} &
  1.0 &
  1.40 &
  \multicolumn{1}{c|}{1.0} &
  \multicolumn{1}{c|}{1.0} &
  1.0 \\ \hline
3 &
  1 &
  \multicolumn{1}{c|}{0.011} &
  \multicolumn{1}{c|}{0.049} &
  \multicolumn{1}{c|}{1.0} &
  \multicolumn{1}{c|}{1.0} &
  \multicolumn{1}{c|}{0.977} &
  1.0 &
  1.43 &
  \multicolumn{1}{c|}{-} &
  \multicolumn{1}{c|}{1.0} &
  1.0 \\ \hline
4 &
  1 &
  \multicolumn{1}{c|}{0.004} &
  \multicolumn{1}{c|}{0.012} &
  \multicolumn{1}{c|}{1.0} &
  \multicolumn{1}{c|}{1.0} &
  \multicolumn{1}{c|}{1.0} &
  1.0 &
  1.19 &
  \multicolumn{1}{c|}{-} &
  \multicolumn{1}{c|}{1.0} &
  1.0 \\ \hline
5 &
  1 &
  \multicolumn{1}{c|}{0.005} &
  \multicolumn{1}{c|}{0.016} &
  \multicolumn{1}{c|}{1.0} &
  \multicolumn{1}{c|}{1.0} &
  \multicolumn{1}{c|}{0.86} &
  1.0 &
  1.59 &
  \multicolumn{1}{c|}{-} &
  \multicolumn{1}{c|}{1.0} &
  0.866 \\ \hline
\textbf{Average} &
  \textbf{} &
  \multicolumn{1}{c|}{\textbf{0.008}} &
  \multicolumn{1}{c|}{\textbf{0.029}} &
  \multicolumn{1}{c|}{\textbf{0.964}} &
  \multicolumn{1}{c|}{\textbf{1.0}} &
  \multicolumn{1}{c|}{\textbf{0.913}} &
  \textbf{1.0} &
  \textbf{1.36} &
  \multicolumn{1}{c|}{\textbf{1.0}} &
  \multicolumn{1}{c|}{\textbf{1.0}} &
  \textbf{0.873} \\ \hline
\end{tabular}%
}
\end{table*}

\begin{table*}[]
\centering
\caption{Table showing the performance of SmartFlow on Passport Registration application.}
\label{tab:results-passport}
\resizebox{\textwidth}{!}{%
\begin{tabular}{|c|c|cccccc|c|ccc|}
\hline
\multirow{3}{*}{\textbf{Layout No.}} &
  \multirow{3}{*}{\textbf{Page No.}} &
  \multicolumn{6}{c|}{\textbf{Accuracy}} &
  \multirow{3}{*}{\textbf{\begin{tabular}[c]{@{}c@{}}Task completion average time\\ (in mins)\end{tabular}}} &
  \multicolumn{3}{c|}{\textbf{Complex Component Accuracy}} \\ \cline{3-8} \cline{10-12} 
 &
   &
  \multicolumn{2}{c|}{\textbf{OCR}} &
  \multicolumn{2}{c|}{\textbf{Layout Mapping}} &
  \multicolumn{1}{c|}{\multirow{2}{*}{\textbf{Filled Data}}} &
  \multirow{2}{*}{\textbf{Request Submission}} &
   &
  \multicolumn{1}{c|}{\multirow{2}{*}{\textbf{Datepicker}}} &
  \multicolumn{1}{c|}{\multirow{2}{*}{\textbf{Dropdown}}} &
  \multirow{2}{*}{\textbf{Radio/Checkbox}} \\ \cline{3-6}
 &
   &
  \multicolumn{1}{c|}{\textbf{CER}} &
  \multicolumn{1}{c|}{\textbf{WER}} &
  \multicolumn{1}{c|}{\textbf{Rule-based}} &
  \multicolumn{1}{c|}{\textbf{Virtual-Grid}} &
  \multicolumn{1}{c|}{} &
   &
   &
  \multicolumn{1}{c|}{} &
  \multicolumn{1}{c|}{} &
   \\ \hline
1 &
  1 &
  \multicolumn{1}{c|}{0.005} &
  \multicolumn{1}{c|}{0.019} &
  \multicolumn{1}{c|}{0.9} &
  \multicolumn{1}{c|}{1.0} &
  \multicolumn{1}{c|}{0.94} &
  1.0 &
  1.51 &
  \multicolumn{1}{c|}{0.8} &
  \multicolumn{1}{c|}{1.0} &
  0.8 \\ \hline
2 &
  1 &
  \multicolumn{1}{c|}{0.013} &
  \multicolumn{1}{c|}{0.062} &
  \multicolumn{1}{c|}{1.0} &
  \multicolumn{1}{c|}{1.0} &
  \multicolumn{1}{c|}{0.957} &
  1.0 &
  1.82 &
  \multicolumn{1}{c|}{1.0} &
  \multicolumn{1}{c|}{1.0} &
  0.8 \\ \hline
3 &
  1 &
  \multicolumn{1}{c|}{0.007} &
  \multicolumn{1}{c|}{0.030} &
  \multicolumn{1}{c|}{0.93} &
  \multicolumn{1}{c|}{0.93} &
  \multicolumn{1}{c|}{1.0} &
  1.0 &
  1.71 &
  \multicolumn{1}{c|}{1.0} &
  \multicolumn{1}{c|}{1.0} &
  1.0 \\ \hline
4 &
  1 &
  \multicolumn{1}{c|}{0.010} &
  \multicolumn{1}{c|}{0.046} &
  \multicolumn{1}{c|}{1.0} &
  \multicolumn{1}{c|}{1.0} &
  \multicolumn{1}{c|}{0.92} &
  1.0 &
  1.58 &
  \multicolumn{1}{c|}{1.0} &
  \multicolumn{1}{c|}{1.0} &
  0.7 \\ \hline
5 &
  1 &
  \multicolumn{1}{c|}{0.011} &
  \multicolumn{1}{c|}{0.034} &
  \multicolumn{1}{c|}{0.81} &
  \multicolumn{1}{c|}{1.0} &
  \multicolumn{1}{c|}{1.0} &
  1.0 &
  1.4 &
  \multicolumn{1}{c|}{1.0} &
  \multicolumn{1}{c|}{1.0} &
  1.0 \\ \hline
\textbf{Average} &
  \textbf{} &
  \multicolumn{1}{c|}{\textbf{0.009}} &
  \multicolumn{1}{c|}{\textbf{0.038}} &
  \multicolumn{1}{c|}{\textbf{0.928}} &
  \multicolumn{1}{c|}{\textbf{0.986}} &
  \multicolumn{1}{c|}{\textbf{0.963}} &
  \textbf{1.0} &
  \textbf{1.604} &
  \multicolumn{1}{c|}{\textbf{0.96}} &
  \multicolumn{1}{c|}{\textbf{1.0}} &
  \textbf{0.86} \\ \hline
\end{tabular}%
}
\end{table*}

\subsection{Future Scope}
\label{sec:future-scope}
\begin{itemize}
\item \textbf{Handling Dynamic Fields}:
We introduce an iterative approach for managing dynamic fields within a website, which differ from static fields in that their appearance depends on the values entered in preceding fields. Our method operates by populating one field at a time and capturing screenshot of the current application page status. The SmartFlow system then conducts visual analysis on the captured screenshot to determine the next field to be filled. This involves extracting all form elements, identifying unfilled fields, and selecting the highest unfilled form element on the page. Subsequently, a layout mapping is established between this selected field and the corresponding user-task request value. PyAutoGUI code is then generated by the Language Model (LLM) to populate the data field. This process continues iteratively as SmartFlow executes the action, extracts the subsequent field to be filled, thereby facilitating dynamic field handling seamlessly.

\item \textbf{Handling Scrollable Forms}:
In the context of scrollable forms, the SmartFlow methodology can be augmented through an iterative procedure involving incremental webpage scrolling by a few pixels. By scrutinizing the frame differential data using vision based techniques between two screenshots, namely the state before and after scrolling, it is possible to identify the initial field requiring input, mirroring the methodology employed for addressing dynamic fields in the SmartFlow. 

\item \textbf{Deep-learning based identification of form fields}
Within the SmartFlow system, Language Models (LLMs) currently rely on the HTML source code of web pages to identify field names and their associated types within the visual interface. However, to enhance efficiency and eliminate the dependency on HTML source code, we propose a novel approach that leverages deep learning-based computer vision techniques for the automatic identification of field names and their types. This approach involves training a deep neural network specifically designed for field detection and classification. To train this neural network, we compile a diverse dataset comprising various web page layouts. Each layout is meticulously annotated to establish ground truth information, which is traditionally derived from the HTML source code. By training the deep neural network on this dataset, we enable it to learn and understand the intricate visual relationships between field types and their corresponding visual representations on the web page. As a result, the trained model becomes proficient in recognizing field names and their types solely based on visual cues, eliminating the need for parsing HTML source code

\end{itemize}

\end{document}